\documentclass[runningheads]{llncs}

\usepackage{amsmath} 
\usepackage{graphicx}
\usepackage[width=122mm,left=12mm,paperwidth=146mm,height=193mm,top=12mm,paperheight=217mm]{geometry}

\usepackage{tikz}
\usepackage{hyperref}       
\usepackage{booktabs}       
\usepackage{amsfonts}       
\usepackage{nicefrac}       
\usepackage{capt-of}

\newcommand\rowincludegraphics[2][]{\raisebox{-0.4\height}{\includegraphics[#1]{#2}}}

\begin{document}
\title{Learning Disentangled Expression Representations from Facial Images}

\author{Marah Halawa\inst{1,2} \thanks{Research work done for a master's thesis in collaboration between TU Berlin and SAP SE.} \and
Manuel W\"{o}llhaf \inst{1} \and
Eduardo Vellasques \inst{2}\and
Urko S\'{a}nchez Sanz\inst{2}\and
Olaf Hellwich \inst{1}}

\authorrunning{M. Halawa et al.}
%
\institute{Computer Vision \& Remote Sensing, Technische Universität Berlin, Berlin, Germany \\
\email{marah.halawa.ite@gmail.com}, \email{\{woellhaf, olaf.hellwich\}@tu-berlin.de} \\
SAP SE, Berlin, Germany \\ \email{\{eduardo.vellasques, urko.sanchez.sanz\}@sap.com}}

%
\maketitle              
\begin{abstract}
Face images are subject to many different factors of variation, especially in unconstrained in-the-wild scenarios. For most tasks involving such images, e.g. expression recognition from video streams, having enough labeled data is prohibitively expensive. One common strategy to tackle such a problem is to learn disentangled representations for the different factors of variation of the observed data using adversarial learning. In this paper, we use a formulation of the adversarial loss to learn disentangled representations for face images. The used model facilitates learning on single-task datasets and improves the state-of-the-art in expression recognition with an accuracy of\textbf{ 60.53\%} on the AffectNet dataset, \textbf{without using any additional data}.

\end{abstract}


\section{Introduction} \label{introduction}
Automated face analysis has many applications~\cite{application}, such as in medical treatment and human-computer interaction.
However, the different factors of variation, such as age, gender, and identity overlap heavily \cite{li2018deep,sawant2019age}. In-the-wild data is almost always occluded due to head-pose or facial attributes like glasses or beard. 
Especially in facial expression recognition the performance of state-of-the-art methods is still unsatisfying. While the application of deep learning has led to superhuman performance in face identity recognition \cite{cao2018vggface2,Taigman14}, expression recognition on in-the-wild datasets is still an unsolved problem \cite{sfewresult1,kim2015hierarchica}.
The efficiency of many data-driven tasks depends on the quality of the data representation \cite{Goodfellow2016a}, and the developments in deep learning in the last decade allow to learn rich task-specific features given a sufficient amount of labelled data \cite{Bengioy2012}.
Due to the essential role these data representations play in determining the model’s overall performance, recent research has paid attention to these representations’ characteristics.
One way to influence the learned features is disentangled representation learning \cite{Bengioy2012,BengioYann}. It aims to improve learned representations by separating the desired information from other factors, or sources of variation, thus obtaining higher levels of abstraction in captured representations and enabling better generalization \cite{Schmidhuber1992,Bengioy2012,BengioYann}. An example model that learns disentangled representations is an autoencoder~\cite{LeCun1987,Hinton1994}, which performs this by reconstructing the data as a linear combination of low-level features \cite{Bengioy2012}.
Recently, \cite{Liu2018} introduced a framework called identity Distilling and Dispelling Autoencoder (D2AE), which disentangles face identity features from other features. Despite the similarity between our approach and theirs, we show that there is no need for a decoder network for disentangling facial expression representations. 
\cite{li2018occlusion} had achieved state-of-the-art on the AffectNet benchmark with 58.78\% accuracy using a CNN that utilizes attention mechanisms and is co-trained on multiple datasets. On the other hand, without additional attention mechanisms, nor additional datasets, our method outperforms theirs on the same benchmark.
In this paper, we investigate the disentangling of expression representations from other factors of variation from face images. We use an encoder-based adversarial method. The main motivation is to tackle the problem of expression recognition in ``in-the-wild'' scenarios. 

\section{Learning Disentangled Expression Representations}\label{method}
\begin{figure*}[t]
\centering
\includegraphics[width=0.5\textwidth, height=0.2\textheight]{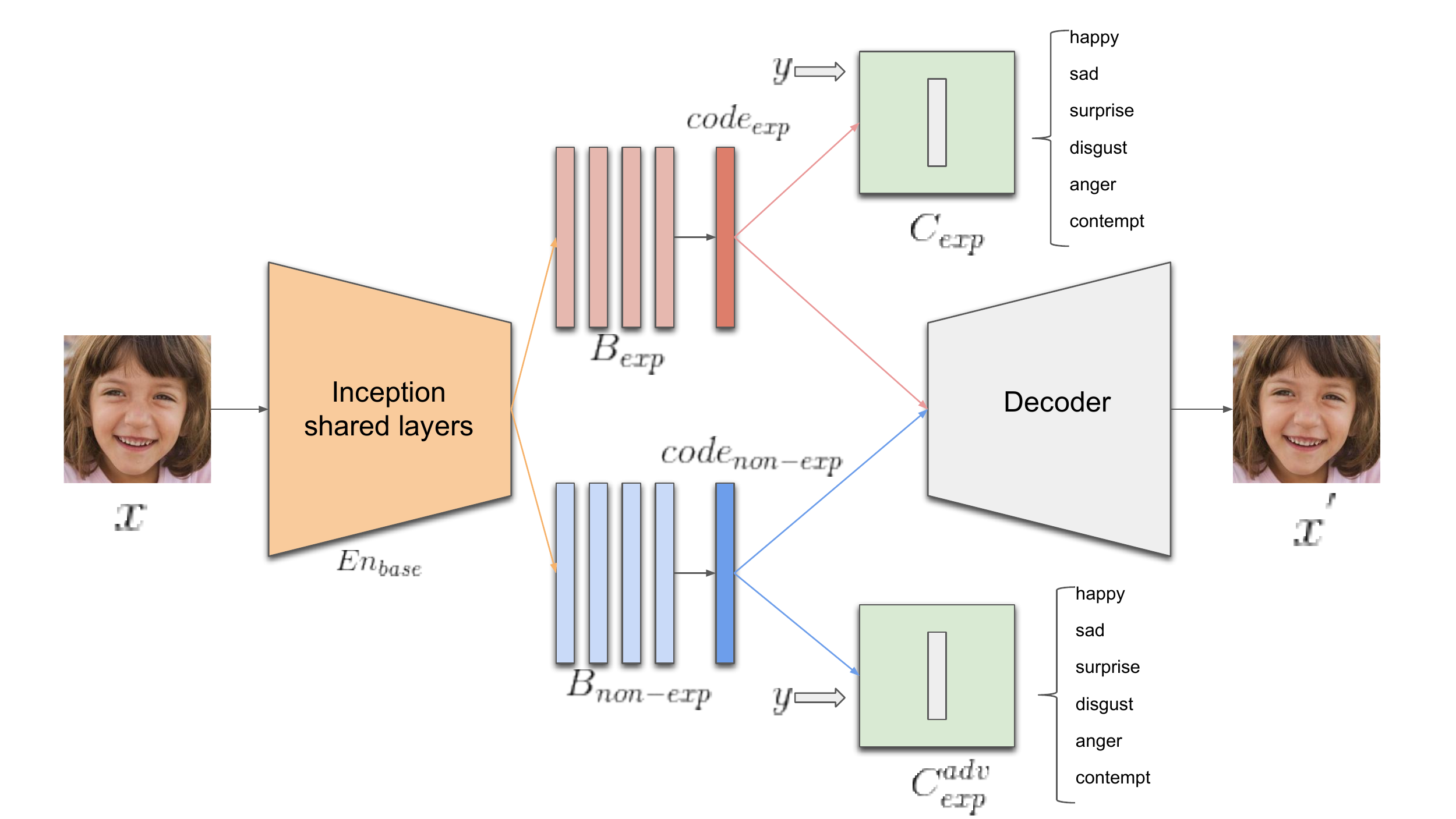}
\caption{Disentangling expression representation from all other non expression features.} \label{fig29} \vspace{-8mm} 
\end{figure*}

The proposed method generates two distinct representations by using two different encoders: the first encodes the facial expression representation from an input image, and the other encodes the representation of other factors of variation.
However, these two encoders share some layers (for shared face features) using a shared encoder $En_{base}$. This shared encoder is followed by the two branches $B_{exp}$, and $B_{non-exp}$, which specialize in encoding expression and non-expression specific features, respectively.
Then, the two encoders are trained in a way that enables a decoder to reconstruct the input image by combining the representations together. 
In order to achieve our goal of disentanglement, we use an adversarial loss that is imposed on non-expression representations, \emph{i.e.} to prevent them from being able to discriminate expressions. This way, we ensure that these representations do not contain any expression features. 
We also introduce another constraint that induces the expression representations to recognize expressions. Thus, the branch $B_{exp}$ is trained in a supervised manner with expression labels to capture expression features using an expression classifier $C_{exp}$. And $B_{non-exp}$ is trained adversarially to disentangle non-expression features using an expression discriminator $C^{adv}_{exp}$. Fig. \ref{fig29} shows our proposed model architecture.
\\
\\
\textbf{Training Process: \label{sec:training1}}
The loss function that guides the training process in the proposed method consists of many terms.
First, a reconstruction loss $\mathcal{L}_{r}$ is imposed on the encoder-decoder, and the two branches $B_{exp}$ and $B_{non-exp}$ to ensure that $code_{exp}$ and $code_{non-exp}$ encode all important features from the input image $x$, respectively. This loss is an $L_2$-norm between the input image $x$ and its reconstruction ${x}'$, and is formally expressed as
$\mathcal{L}_{r} = \left \| x - De(code_{exp} \oplus code_{non-exp}) \right \|^2$.
The second loss is the expression classification loss $\mathcal{L}_{c-exp}$. We use the cross-entropy loss to train this classifier, as shown in eq.\ref{classify-exp}. The optimization over $\mathcal{L}_{exp}$ updates the shared encoder layers $En_{base}$, the expression branch layers $B_{exp}$, and the expression classifier layers $C_{exp}$.
\begin{equation} \label{classify-exp}
-\: \mathcal{L}_{exp} = \sum_{i=1}^{N_{exp}}\; y_{i} \;\log(C_{exp}(code_{exp}))
\end{equation}
where $N_{exp}$ is the number of emotions.
Finally, we introduce the adversarial loss $\mathcal{L}_{adv}$ that is imposed on the output of the expression classifier $C^{adv}_{exp}$, which ensures that $code_{non-exp}$ does not contain any expression features.
$\mathcal{L}_{adv}$, as any adversarial loss, consists of two parts: The first is $\mathcal{L}^{adv}_{exp}$ that is imposed on the output of the classifier $C^{adv}_{exp}$, which classifies $code_{non-exp}$ into one of the expressions. The optimization over $\mathcal{L}^{adv}_{exp}$ updates only the expression classifier layers $C^{adv}_{exp}$. 
The second part $\mathcal{L}^{adv}_{En}$, tries to fool the expression classifier $C^{adv}_{exp}$ by maximizing the uncertainty of the classifier's output for input images. Therefore, the optimization over $\mathcal{L}^{adv}_{En}$ updates the shared encoder layers $En_{base}$, and non-expression branch layers $B_{non-exp}$. The formulation of these losses is as follows:
\begin{equation} \label{adv_gan}
\begin{split}
- \mathcal{L}^{adv}_{exp} &= \sum_{i=1}^{N_{exp}} y_{i} \log(C^{adv}_{exp}(code_{non-exp})) \\
- \mathcal{L}^{adv}_{En} &= \frac{1}{N_{exp}}  \sum_{i=1}^{N_{exp}} \log(C^{adv}_{exp}(code_{non-exp})) \\
\mathcal{L}_{adv} &= \mathcal{L}^{adv}_{exp} + \mathcal{L}^{adv}_{En} 
\end{split}    
\end{equation}

In order to train the proposed framework, we minimize the final objective function which is the weighted sum of all the above losses, and is given by $\mathcal{L}_{final}$ as follows:
\begin{equation} \label{final1}
\mathcal{L}_{final} =  \beta_{1}\;\mathcal{L}_{r} + \beta_{2}\;\mathcal{L}_{exp} + \beta_{3}\;\mathcal{L}_{adv} 
\end{equation}

Where $\beta_{1}$, $\beta_{2}$, $\beta_{3}$, are hyper-parameters to control the contribution of each loss. We should mention that we can optimize $\mathcal{L}_{r}$, $\mathcal{L}_{exp}$, and $\mathcal{L}^{adv}_{En}$ jointly. However, $\mathcal{L}^{adv}_{exp}$, is optimized independently from the above losses, and it updates $C^{adv}_{exp}$ while the other components of the model remain constant.

\section{Experiments} \label{experiments}

\textbf{Experimental setup} \label{experiment-exp-nexp-setup}
For the following experiments, we use the model in Fig. \ref{fig29}, where each expression branch $B_{exp}$ and non-expression branch $B_{non-exp}$ consists of four Conv. layers, with leaky-ReLU activation functions. 
The decoder $De$ consists of six De-conv. layers with leaky-ReLU activation functions, except for the last layer with a Sigmoid activation. 
The shared encoder $En_{base}$ is an Inception-ResNet-V1 \cite{szegedy2017inception} encoder, pre-trained on VGGFace2 dataset \cite{cao2018vggface2}. Both $C_{exp}$ and $C^{adv}_{exp}$ are linear classifiers.
The values of the hyper-parameters in eq.\ref{final1} are $\beta_{1} = 0$, $\beta_{2} = \beta_{3} = 1$. Thus, there is no effect for the reconstruction loss, except for the experiment with an ablation study for the influence of the reconstruction loss.
For all our experiments, we report the validation accuracy of $C_{exp}$ and $C^{adv}_{exp}$. However, when comparing with methods from the literature, we only provide results for $C_{exp}$, since other methods do not rely on an adversarial loss. The baseline model architecture is identical to the above model, only without the adversarial components. This means that the baseline consists of an encoder similar to $En_{base}$ in Fig. \ref{fig29}. This encoder is followed by four Conv. layers similar to $B_{exp}$. Then, the learned representation is fed as an input to the linear classifier $C_{exp}$ similar to $C_{exp}$ in Fig. \ref{fig29} as well. 
\\
\\
\textbf{Results}
\label{experiment-exp-nexp-results1}
The first row in Table \ref{tab:results5} shows that we outperform the baseline on the AffectNet dataset, confirming that our method provides a performance boost on the task of FER. In Table \ref{tab:results5} we also compare to methods from literature, and report the classification accuracy. We obtain the results for these methods from \cite{li2018occlusion}. The first method (VGG16) is a simple classifier that uses the VGG16 architecture \cite{simonyan2014very} trained on a mix of RAF \cite{li2018reliable} and AffectNet \cite{mollahosseini2017affectnet} datasets. In the IPA2LT \cite{Zeng_2018_ECCV} framework, the samples are assigned to multiple labels. This method is also trained on a combination of RAF and AffectNet, and on 1.2 million unlabelled face images from AffectNet and Microsoft Bing. We also compare to gACNN \cite{li2018occlusion}, which relies on attention mechanisms in this task, and to DLP-CNN \cite{li2018reliable}, which simulates how attention mechanisms work using a CNN. It is noteworthy that our method outperforms all the above methods. It is worth mentioning that there are few works that achieve better accuracy on AffectNet by using additional datasets or multi-modal data \cite{cake} \cite{NIPS2019} \cite{steven2}.

The ablation study for the effect of the reconstruction task is shown in Table \ref{tab:results-en-de-loss}, which shows the classification accuracy of $C_{exp}$ and $C^{adv}_{exp}$ on AffectNet dataset when we set the hyper-parameter $\beta_{1}$ in eq.\ref{final1}. 
The results in the table confirm that the reconstruction task has a negative effect on disentangling, e.g. when $\beta_{1}=1$ or $\beta_{1}=0.001$ both representations $code_{exp}$ and $code_{non-exp}$ have a similar ability in encoding expression features, which means less disentangling performance. However, we achieve the best performance when 
there is no effect for the reconstruction loss $\beta_{1} = 0$. Higher $C_{exp}$ accuracy indicates better classification results, however, lower $C^{adv}_{exp}$ accuracy indicate better disentangling behavior.
In order to evaluate the quality of the learned representations on other datasets, we chose randomly a few images from the CASIA dataset \cite{casia}. Table \ref{tab:results-cos-similar} shows the cosine-similarity between the representations of these images, and shows that images with the same expressions have more similar representations regardless of their identities.

\begin{table}
\parbox[t]{.45\linewidth}{
    \centering         \small
    \label{fig35}
    \begin{minipage}[t]{0.45\textwidth}
    \centering
    \includegraphics[width=0.8\textwidth, height=0.2\textheight]{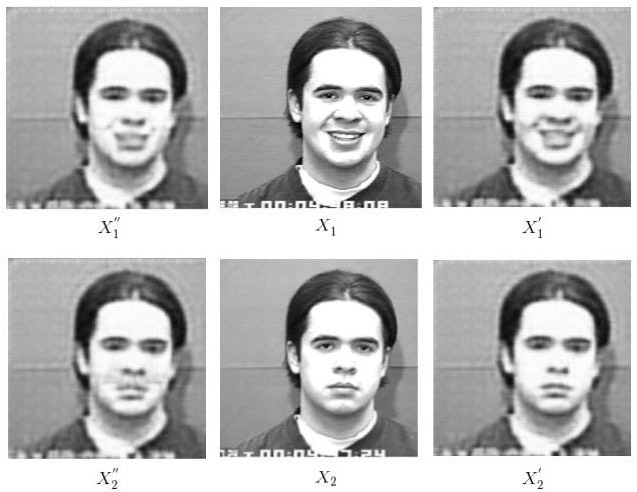}
    \captionof{figure}{\label{fig89}Synthesize images from happy and neutral images of the same identity}
    \end{minipage}
}
\hfill
\parbox[t]{.45\linewidth}{
    \centering         \small
    \caption{Comparison to results from literature on AffectNet dataset}\label{tab:results5}\vspace{-5mm}
    \begin{tabular}[t]{ c c } \toprule
    Approach & Test Accuracy \\
    \hline
    Baseline & 56\%\\
    \hline
    VGG16 \cite{simonyan2014very} & 51.11\% \\
    DLP-CNN \cite{li2018reliable} & 54.47\% \\
    IPA2LT \cite{Zeng_2018_ECCV}  & 57.31\% \\
    gACNN \cite{li2018occlusion} & 58.78\%\\
    \hline
    Proposed method & \textbf{60.53\%}  \\
    \hline
    \end{tabular}
}
\vspace{0.5cm} \\
\parbox[t]{.45\linewidth}{
    \centering         \small
    \caption{Studying the effect of the reconstruction loss on the disentangling performance}\label{tab:results-en-de-loss}\vspace{-5mm}
    \begin{tabular}[t]{ c c c } \toprule
    Hyper-parameter & $C_{exp}$ Acc. & $C^{adv}_{exp}$ Acc. \\
    \hline
    $beta_1=1.0$& 51\%  & 46\% \\
    $beta_1=0.001$& 56\% & 53\% \\
    $beta_1=0.0$& \textbf{60.53\%} & \textbf{16\%}\\
    \hline
    \end{tabular}
}
\hfill
\parbox[t]{.45\linewidth}{
    \centering         \small
    \caption{Cos-Similarity between two expression representations learned by the proposed disentangling method}\label{tab:results-cos-similar} \vspace{-5mm}
    \begin{tabular}[t]{ c c c } \toprule
    Im-1 & Im-2 & Cos-sim.(Im-1,Im-2) \\ \hline
         \rowincludegraphics[scale=0.1]{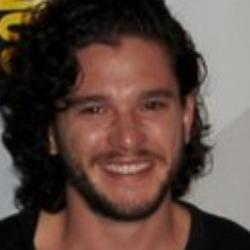} & \rowincludegraphics[scale=0.1]{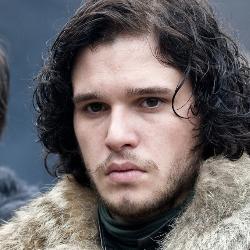} & 0.7877541 \\ \hline
         \rowincludegraphics[scale=0.1]{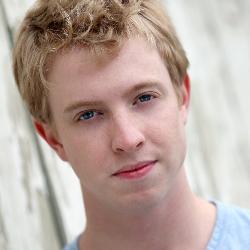} & \rowincludegraphics[scale=0.1]{fig/n_jon.jpg}& 0.83162177 \\ \hline
    \end{tabular}
}\vspace{-1cm} \\
\end{table}

\section{Conclusion and Future Work} \label{conclusion}
This work aimed to improve the performance of facial expression recognition in the wild. To achieve that, we proposed applying an adversarial method to disentangle expression representations from other factors of variation in facial images.
We improve state-of-the-art results \cite{li2018occlusion} on the AffectNet benchmark to \textbf{60.53\%} accuracy, \textbf{without using any additional data}. We also show that the reconstruction constraint is detrimental to the disentangling performance. 

Our work boosts the performance on the FER task, yet we see a room for improvement when disentangling \emph{known} causal factors of variation. This is in line with the findings of \cite{Locatello2018} too. Therefore, in a preliminary set of experiments, we disentangle the identity representation from the expression representation and vise versa, by imposing additional adversarial loss on expression representation similar to eq.\ref{adv_gan}, using identity labels instead. 
The qualitative results in Fig.\ref{fig89} show a clear disentangling between expression and identity representations on the CK+ dataset \cite{lucey2010extended}. Where $X_1$ and $X_2$ are the original images, and $X^{'}_1$ is obtained by decoding both representations of $X_1$, and so is $X^{'}_2$ from $X_2$. $X_{1}^{''}$, and $X_{2}^{''}$ are synthesized by swapping the expression and identity representations of $X_1$ and $X_2$. Therefore, as a future work, we plan to examine disentangling multiple known factors of variation and generalize this work to multiple datasets.

\bibliographystyle{splncs04}
\bibliography{references}

\end{document}